\documentclass[preprint,12pt]{elsarticle}

\usepackage{amssymb}
\usepackage{amsmath}
\usepackage{booktabs}
\usepackage{multirow}

\begin{document}

\begin{frontmatter}



\title{A novel scheme for word-level based offline text-independent writer identification}


\author[inst1]{Vineet Kumar}

\author[inst1]{Suresh Sundaram}
\affiliation[inst1]{organization={Department of Electronics and electrical Engineering},
            addressline={Indian Institute of Technology Guwahati}, 
            city={Guwahati},
            postcode={781039}, 
            state={Assam},
            country={India}}



\begin{abstract}
In this   paper, we  formulate a  novel  scheme to identify the authorship of a document based on  handwritten input word  images of an  individual. Our approach is text-independent and does not place any restrictions on the size of the input word images  under consideration. To begin with, we adapt the SIFT algorithm to extract multiple key point regions at various levels of abstraction (comprising allograph, character or combination of characters). These key point regions are then passed through a  trained CNN network to generate feature maps corresponding to a convolution layer.  However, owing to the   scale  corresponding to the SIFT key points, the size of a generated feature map may differ. As an alleviation to this issue,  we propose to consider the histogram of gradients  on each of the generated  feature maps,  thereby producing a fixed representation.

Typically, in a CNN,  the  number of filters of each convolution block increase  depending on the depth of the network. Accordingly, the extraction of histogram features on each of the convolution feature map increase the dimension as well as the computational load. To address this aspect, we propose a entropy-based measure to learn `saliency' values of the feature maps of a particular CNN layer during the training phase of our algorithm. The  measure, as such is derived from the application of sparse PCA on the histogram of gradient features.

The efficacy of our  system has been demonstrated on  two publicly available databases namely CVL and IAM. We empirically show that the results obtained are promising when compared with previous works.
\end{abstract}




\end{frontmatter}


\section{Introduction}
\label{sec:Introduction}
Biometrics refers to the automatic identification or verification of a person by utilizing either biological \cite{Xu:2018} or behavioral \cite{Chi:2018} features. Under the  purview of behavioral bio metrics, writer identification and verification  have gained popularity as a topic of  research in recent times with  varied applications in the field of forensic analysis \cite{Fernandez:2010}, historical document analysis \cite{Bulacu:2007}, \cite{He:2014} and security \cite{Faundez:2020}. \par 

Writer identification is related to finding the authorship of a given unknown document from a set of reference documents stored in the database based on learned features / descriptors. Depending on the contents of the query and reference document, the writer identification technique can be divided into text-dependent or text-independent approaches. In former,  the author needs to reproduce a sample of given text based on which the identity is established. Signature verification falls under the paradigm of a text-dependent system \cite{Sharma:2017}. Contrast to this,  text-independent approach, on the other hand, identifies the writer regardless of the handwritten  material, as  long as sufficient number of samples are available to extract robust features. 

Depending on the handwriting acquisition modality being employed, writer identification systems can be further divided into  two paradigms, namely  offline and online. In  the online based approach, spatio-temporal information (in  the form of $x$ and $y$ coordinates) along with other features (such as pressure, azimuth and inclination) is recorded by the data acquisition system, which are then subsequently employed to characterize a writer.  Contrary to this, in offline techniques, handwritten data is presented in the form of a scanned image, following which methodologies are suggested to extract various allographic features of the image to identify the writer. In this paper, we will mainly focus  our research towards the development of an offline text-independent writer identification. 

Based on the  literature  of offline text-independent writer identification, the works  can be divided into  one of the categories, namely  texture, shape and deep learning-based approach \cite{Xiong:2017}. The systems relying on texture analysis consider each handwritten input sample as a different texture and accordingly extract a set of features for description in the frequency or spatial domain respectively. Said \textit{et al.} \cite{SAID:2000} proposed a multi-channel 2-D Gabor filter-based approach for texture analysis by deriving attributes at different frequencies and orientations. Bertolini \textit{et al.} \cite{Bertolini:2013} used Local phase quantization (LQP) \cite{Heikkila:2009} features extracted from 2-D short term Fourier transform (STFT) for writer identification. 

Contrary to frequency based methods for the description of texture, spatial techniques treat handwriting as a combination of edges and contours, with their statistical distributions being employed for writer description. Djeddi \textit{et al.} \cite{Djeddi:2013} used run-length features based on the gray level run matrix (GLRM) to describe the handwriting characteristics. Balacu \textit{et al.} \cite{Bulacu:2003} proposed edge based attributes namely: edge-hinge, and edge-direction features to ascertain the individuality of handwriting. Hannad\textit{et al.} \cite{Hannad:2016} divided the handwriting image into small fragments upon which a set of features namely Local Binary Pattern (LBP), Local Ternary Pattern (LTP), and Local Phase Quantization (LPQ) were extracted for writer identification. Wu \textit{et al.} \cite{Wu:2014} employed the SIFT features and its scale and orientation information  from the handwriting samples to characterize a writer.  Likewise, in Faraz \textit{et al.} \cite{Khan:2018}, the authors used a combination of  SIFT and RootSIFT to construct a set of Gaussian mixture models namely similarity GMM (SGMM) and Dissimilarity GMM (DGMM). In an essence, these models aimed at capturing the intra-class similarity and inter-class dissimilarity between same and different writers of the enrolled database. Last but not least, we also make a mention of  oriented basic image features \cite{Newell:2014} that have also been employed for the purpose of writer identification.\par

The prominent works on shape-based approach divide the handwriting into a group of segmented shapes and use the statistical description of their features to characterize the handwriting. Schomaker \textit{et al.} \cite{Schomaker:2004} constructed a code-book based on connected-component contours ($CO^{3}$) to describe the shape of frequently  occurring allograph for a set of enrolled writers. Thereafter,  this information is used to generate a histogram, that is specific  for each individual writer. Siddiqi \textit{et al.} \cite{Siddiqi:2010} used a combination of shape and texture features in the form of code-book and curvature to represent the  features of a writer. He \textit{et al.} \cite{Schomaker:2015} proposed a junction detection based approach for identification, wherein they utilized the information of various types of junctions created by an author to differentiate against other writers. Akram \textit{et al.} \cite{Akram:2019}  used Harris corner detection  to extract key points in the form of junctions and corners from a handwritten image. These keypoints are then used to construct a codebook based on which writer identification is performed. \par

The advancements of machine learning in the recent decade have led to the prominence of deep learning based methods  in the area of computer vision related applications. Compared to traditional hand crafted features,  these techniques learn data dependent characteristics automatically from the training samples, thus resulting in higher performance. The pioneering attempt using deep learning for writer identification was introduced by Fiel \textit{et al.} \cite{Fiel:2015}.  Here, the authors trained a CNN model from writer samples and extracted feature from the penultimate layer to construct a  feature vector.  In other related works,  Xing \textit{et.al} in \cite{Xing:2016} proposed a multi-stream CNN network called DeepWriter,  wherein two adjacent segmented writer patches are passed trough the network and their combined features  utilized for identification. Christlein \textit{et.al}  \cite{christlein:2017(b)} used a clustering based approach to train the weights of the convolution network in an unsupervised manner, thereby enabling the system to learn robust local features. Inspired from the idea of Droupout, Yang \textit{et.al} in \cite{Yang:2016} introduced the concept of Drop Stroke, that randomly removes some of the strokes from the input image. The resulting  modified image along with the original was used to train the weights of the  Convolution network. An empirical study of the effect of individual convolution layers  on the writer identification  rate was performed in the work  \cite{Rehman:2019} . Sulaiman \textit{et.al} in \cite{Sulaiman:2019} employed  the idea of combining  deep and hand-crafted features to represent a writer descriptor. \par

Majority of the aforementioned  works use page or text-line level data to learn writer features. However, in applications such as forensic examination, often a situation arises where a decision needs to be made based on  analyzing very small amount of handwritten text such as  words. This task poses a challenge as the writer-related style information is restricted in contrast to page / text-line level input. Nonetheless, the prominent work in this direction  is \cite{He:2019},  where the authors  employ a multi task framework  to enhance writer related information by incorporating attributes learnt in the auxiliary task along with writer features from the main task. In a  subsequent work \cite{He:2020}, the same authors proposed a deep neural network model called Fragnet, that uses a feature pyramid to extract important writer specific features based on which identification is made. The recent exploration is that of  \cite{He:2021},  where  the authors  exploit the spatial relationship between the sequence of fragments  by a  recurrent neural network (RNN).

\section{RESEARCH FRAMEWORK}
 In this paper, we propose a HOG based writer identification system that uses CNN network to identify a writer using a word image. Our work draws inspiration from Discriminatively Trained part-Based Models \cite{Felzenszwalb:2009} which treats an object as a collection of multiple segments that are trained separately and combined to recognize an object. Analogous to that, we treat a word as a combination of multiple small segments containing information in the form of patterns. These patterns are trained separately following which they are combined to get the identity of a writer. \par

We utilise the SIFT method \cite{Lowe:2004} to generate numerous relevant segments from a word image. This algorithm uses a multi-scale analysis to extract keypoints at different levels of abstraction (comprising  allograph,  character, or combination of characters). The detected SIFT keypoints is then passed through a CNN network  that is pre-trained on the English alphabets of the EMNIST dataset.

In the present work, corresponding to each fragment, the output feature map obtained from the specific convolution layer  (selected experimentally) is used for feature representation \footnote{This is contrary to works wherein the output of the penultimate layer are used as features.}. The size of the feature maps relative to the selected convolution layer varies depending on the scale at which the SIFT keypoint patches are extracted. In order to achieve a fixed  dimension representation from the different size  feature maps, a Histogram Of Gradient (HOG) approach is proposed. Such a gradient  based  representation aids in preserving the spatial relationship between different regions of a feature map. 

Owing to the fact that each layer in a CNN  generates a set of feature maps,  we observe that extraction and saving of the HOG feature vectors  corresponding to each map   increases the computational load. To alleviate this issue, we propose a entropy-based measure to learn `saliency' values of the feature maps of a particular CNN layer during the training phase of our algorithm. The  obtained saliency values of feature maps of a layer in CNN are then subsequently employed for combining them to a single  representation.  This idea, as we shall see later, brings down the  the computational cost without   compromising on the loss of information that can be possibly incurred as a result of combination. 

 The resulting combined  map  (after incorporation of saliency values) is used to generate a HOG based feature vector for each writer, following  which scores are obtained from  a SVM classifier.  It is worth noting here that  the SVM classifier takes as input, the HOG representation of the combined feature maps  for each  of the individual  SIFT fragments of  a word image and outputs a score.  These are then accumulated across all the fragments of the input word   and employed for  establishing the  identity of the writer.  \par

\begin{figure*}[t] 
\centering
\includegraphics[width=\textwidth]{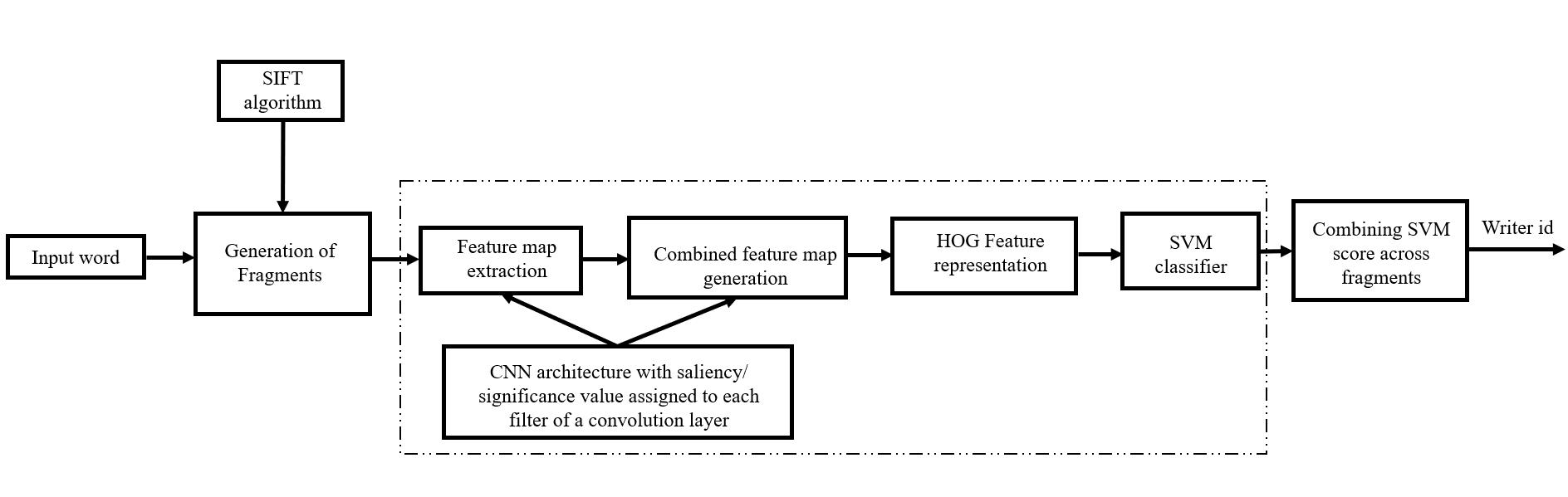}
\caption{Pictorial Overveiw of our proposed system. The blocks enclosed in dotted lines represent the operations that are performed on each writer fragments.}
 \label{fig:outline}
\end{figure*}


Fig.\ref{fig:outline} summarizes the overview of our proposed algorithm. As a reiteration of the preceding discussion, the input  word image is divided into a number of fragments of different sizes.  Thereafter, on each of these individual  fragments the following set of operation is performed:
\begin{itemize}
    \item Generation of feature maps in each  convolution layer of the CNN network.
    \item Combination of feature maps of a layer into a single representation
    \item HOG  feature extraction on the combined feature map to get a fixed dimension  description of a fragment. 
    \item Generation of scores from a trained SVM. 
\end{itemize}
The SVM  scores across fragments are accumulated to get the identity of the writer.



The rest of the paper is organized as follows. In Section \ref{sec_3} we describe the process of fragment generation using the SIFT keypoint detector. This is followed by a detailed discussion of the CNN block used for feature extraction  in Section \ref{sec_4}. In Section \ref{sec_5},  we provide details of the HOG approach  for   obtaining fixed size representation of the feature maps. The methodology proposed for associating saliency values to each layer of a  convolution block is discussed at length in Section 6. In Section \ref{sec_7} the  incorporation of the saliency  values for the generation of a combined feature map is  elaborated. 

With regards to the experiments, the   dataset used for evaluating our algorithm  is presented in Section \ref{sec_8} together with the training and testing protocol. In Section \ref{sec_9} we evaluate the efficacy of our proposed writer identification algorithm and compare its performance with other prior works.  Finally, we summarize our paper in section \ref{sec_10}.    

\section{Fragment generation step}\label{sec_3}

Locating important features  across images is a common task prevalent in various domains of computer vision.
Many deep learning based techniques in the field of writer identification use dense sampling for extracting multiple handwriting blocks from a given input sample. However,  prior studies \cite{zhang:2003,Tan:2010} have suggested that all the characters in a handwritten image need not be equally informative in characterizing a writer. Secondly,  dense sampling may result in important writer features getting split across several windows  thus leading to a loss of information. As a circumvention to this, important attributes in an image (such as corners) are identified and the region around them extracted  as fragments using a pre-determined window size.  As such, these fragments   are more likely to discriminate a writer effectively as compared to the image patches extracted using dense sampling.\par

A simple key-point detector such as Harris corner   works effectively, when the overall features of images under consideration are similar (in terms of scale and orientation). However, in case of a handwritten document, a writer may create a text that  can differ in scale, whereby   a fixed sized window may not be always effective in capturing different sized features. This observation motivated us to  employ the  popular scale invariant based interest point detector   (SIFT)   for extracting the fragments  around  each of the  key points detected in the image \cite{Wu:2014,Khan:2018,Christlein:2017}. The  SIFT algorithm is divided into four stages: In the first stage an input image is broken into a Gaussian pyramid, the size of image at each successive level of the pyramid (called an octave) is half of the preceding one.  In the second stage, the image at each octave is used to generate a Difference of Gaussian (DoG) images, by convolving them with a series of Gaussian filer of different variances. In the  third stage, the difference of Gaussian (DoG) images generated in the above step is used to detect stable keypoints, following which the location, scale and orientation of these keypoints are computed. Finally, a fixed dimesional feature descriptor is constructed to represent the image features using Histogram of Oriented Gradient (HOG).   \par

\begin{figure}[t] 
\centering
\includegraphics[width=\textwidth]{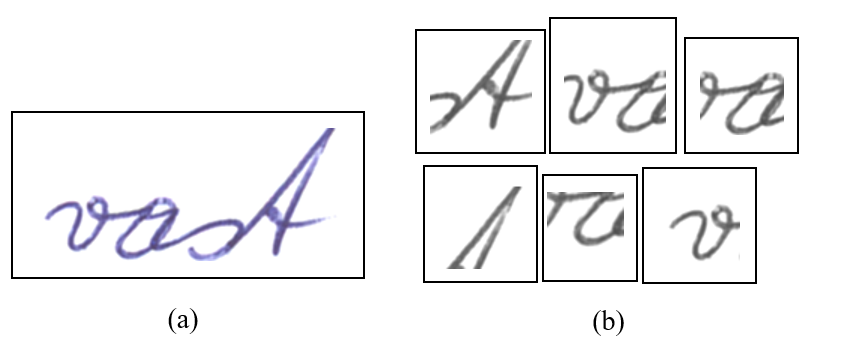}
\caption{Example of keypoint detection using SIFT. (a) segmented Input word image, (b) detected Keypoint}
 \label{SIFT}
\end{figure}

In our implementation of the SIFT algorithm, we use the scale and orientation information of the detected keypoints to extract an image patch around that location. Some of the keypoints extracted using SIFT algorithm on input word image is shown with the help of Fig. \ref{SIFT}.  

\section{CNN feature descriptor}\label{sec_4}

A CNN architecture consists of a combination of several convolution blocks each having a set of filters, whose number increases progressively as we go deeper into the network. Each set of filters in a convolution block, extracts feature from the preceding blocks . These input features are extracted by convolving the learned filter weights over the preceding input layer, with each filter sharing its weight across the input. The filter weight is learned during training stage by tuning it continuously to optimize the loss function. Convolution operation in a CNN can be mathematically represented as:
\begin{equation}
    x_{j}^{l}=f(\sum_{i\in M_{j}}x_{i}^{l-1} \times k_{ij}^{l}+b_{j}^{l})
\end{equation}
Here, $x_{j}^{l}$ represents the $j^{th}$ feature map of layer $l$, $f$ represents the non-linear activation function, $M_j$ represents a section of the input feature map, $k$ represents the convolution filter kernel, $l$ represents the depth of the convolution layer and $b$ represents the bias term.\par

\begin{figure}[t] 
\centering
\includegraphics[width=\textwidth]{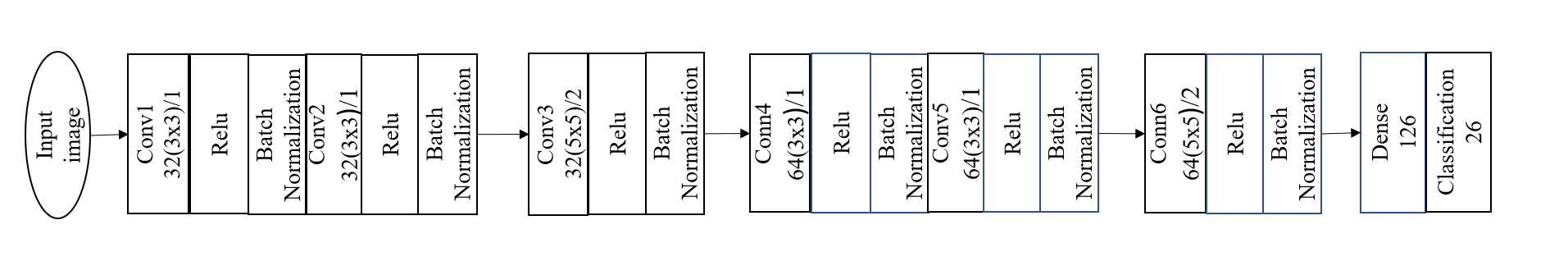}
\caption{CNN network used for training English character on EMNIST dataset \cite{Cohen:2017}. }
\label{CNN_2}
\end{figure}

Most of the CNN based algorithms in the field of writer identification use images patches of a fixed predefined size  extracted at text-line/ document level to train a convolution network for classifying writers  \cite{Fiel:2015,Tang:2016,Xing:2016,Sulaiman:2019}. In the process of designing these type of writer dependent convolution network we are faced with the following challenges: 
\begin{itemize}
    \item Performance of such a model is dependent on quality of database. To properly train a convolution network, there must be enough training examples to capture a variety of writing patterns. Furthermore, the required amount of training example increases as the model gets bigger. 
    
    \item In a convolution network trained to identify multiple classes adding or removing a class posses a problem, as, the network needs to be retrained on the whole dataset. 
    
    \item  A writer dependent network is not able to generalize writer feature across different dataset, as such the performance of the system decreases when the convolution network trained on a particular dataset is used to test writer samples form a different dataset \cite{He:2020}. 
\end{itemize}
Due to the above mentioned reason taking inspiration form the field of unsupervised learning \cite{dosovitskiy:2014}, we use a writer-independent dataset to train the convolution network enabling it to learn diverse set of features. Our CNN network is trained on EMNIST dataset \cite{Cohen:2017} consisting of a combination of handwritten English alphabets and digits of which, we only use handwritten English alphabets.  Our network consists of six convolution block, with each convolution block succeed by Rectified Linear Unit (ReLU) and batch normalization layer respectively. The block diagram of our Convolution network is shown in fig. \ref{CNN_2}. Each convolution layer has been assigned a number from 1 to 6, which is followed by the number of filter and the size of filter \footnote{conv1,$32(3\times3)/1$ signifies convolution 1 layer consists of 32 filters each of kernel size $(3\times 3)$ with stride of 1 }. In our implementation, we have replaced the max-pooling operation with stride convolution, as it helps in better visualization of lower
layers features of the network without compromising on accuracy \cite{Springenberg:2014}.

\section{HOG feature representation}\label{sec_5}
Taking into consideration the size of the feature map obtained by passing the variable sized image fragment through the convolution block, there is requirement to transform the output into a fixed sized representation. To transform the feature map into a fixed-size representation we use Histogram of Oriented Gradients (HOG) \cite{Dalal:2005} based feature representation. In, this section we provide the details of our modified HOG representation.\par

Histogram of Oriented Gradients (HOG) is widely used in the field of object detection to extract features from an image. This algorithm takes into account the gradient magnitude and orientation information of an image to encode its features. This algorithm preserves the spatial information besides being invariant to small geometric transformations. Our HOG based feature representation is obtained by implementing the  following steps in sequence, this implementation draws motivation from SIFT feature representation:
\begin{itemize}
\itemsep.8em 
    \item An input image is first divided into a fixed number of sub-block, with each sub-block containing a certain number of pixels, which is adjusted based on the size of the feature map.
    \begin{equation}
    \begin{split}
        r_{cell} &=\Bigl\lceil \frac{M}{m} \Bigl\rceil \\
        \\
        c_{cell} &= \Bigl\lceil \frac{N}{n} \Bigl\rceil
    \end{split}
    \end{equation}
    Here, $m$ and $n$ represents the number of image grids along the row and column based on which the size of row and column cell $(r_{cell},c_{cell})$ is adjusted.
    
    \item Resulting $(m\times n)$ sub-image blocks are grouped into $b$ number of block, with each block containing $t$ number of sub-image each of size ($r_{cell} \times c_{cell}$).

    \item   Gradient  $G_{x}$ and $G_{y}$ along $x$ and $y$ direction calculated for the whole image, which is used to obtain magnitude and orientation of each pixel using equation (\ref{grad}).
    
    \begin{equation} 
        \begin{split} \label{grad}
        \textbf{Magnitude} &=\sqrt{G_x^2+G_y^{2}}\\
        \\
        \textbf{Orientation} &=atan\Big(\frac{G_y}{G_x}\Big)
        \end{split}
    \end{equation}
    
    \item Magnitude and orientation information obtained in the above step is used to construct a histogram relative to each sub-block of image. The horizontal axis of the histogram represents the orientation information discretized into a fixed set of bins in  multiples of $\phi=\Bigl\lceil \frac{360}{k} \Bigl\rceil$ degrees, where $k$ refers to number of bins in the Histogram. Location ($l$) of bin to be voted is selected based on orientation  $(l=\Bigl\lceil \frac{Orientation}{\phi} \Bigl\rceil)$, and the vote (the value that goes into the bin) is selected based on the magnitude.
    
    \item Corresponding to each of the $x^{th}$ sub-images $(1\leq x\leq t)$ in the $y^{th}$ block $(1\leq y\leq b)$ histogram $S_{xy}$ is constructed as explained above. Each of these histograms corresponding to a given block $y$ $(1\leq y\leq b)$ is concatenated to get an overall $k\times t$ dimensional histogram representation for a given block. Finally the histogram for each of these $t$ blocks are stacked together  to generate a final $k\times t\times b$ dimensional feature vector.
     \begin{equation}
        \begin{split}\label{comb_hog}
        F_y &=[S_{1y}\; S_{2y}\; ....\; S_{xy}] \quad 1\leq x \leq t\\
        \\
       \textbf{FV} &=\frac{[F_{1}\; F_{2}\; ... \;F_{y}]}{||[F_{1}\; F_{2}\; ... \;F_{y}]||_2} \quad 1\leq y \leq b 
       \end{split}
     \end{equation}
\end{itemize} 

\section{Determination of Saliency value for convolution filters}\label{sec_6}

The HOG feature representation learned for the output SIFT keypoint fragments of a Convolution network needs to be stored for training the writer classifier in subsequent stages. Storing HoG features corresponding to all the filters of a convolution layer is a memory intensive and computationally expensive task. A solution to this problem involves combining the output feature map of the convolution layer to summarize the extracted information of a convolution block. The simplest way of achieving this objective is by assigning equal weights to the output feature map of a convolution layer.However, considering the fact that all the filters of a convolution layer are not equally informative, we propose a entropy-based method, which assign a saliency value to each filters of a convolution layer based on the amount of information they contain. In this section, we explain in detail our proposed entropy-based method for assigning saliency score to a convolution layer filters. \par  

For calculating saliency score, we select a set of $\textbf{W}$ writes form the database listed in Table\ref{datset}. Corresponding to each of these selected writers fragments, HOG feature is extracted relative to each feature map output of a convolution layer for which saliency value need to be determined. On these HOG features following operations are performed:
\begin{enumerate}
    \item Extraction of principal components from HOG feature and constructing a histogram using it.
    \item Computation of entropy by utilizing the histogram generated in the above step and using it to assign a  saliency value to each convolution layer filter.
\end{enumerate}
In the following subsection we explain in details each of the above mentioned steps.
\subsection{Histogram generation using Sparse Principal component analysis}
The first step in our objective of assigning a saliency value to a filter, involves projecting $W$ writer HoG features onto a common subspace (as done in traditional $K$-means clustering algorithm). This projection matrix  represents the significance of each projected component in the subspace.\par

For effectively analyzing a data Principle Component Analysis (PCA) is one of the most commonly used technique. It involves transforming the original variables into a set of new variables called principle component (PC). These set of principle component are uncorrelated and represent degree of variation along the data. Thus, PCA makes model interpretation and visualization easier by localizing information in only few components, however these components are constructed using all input variable making them difficult to interpret.\par

Sparse PCA \cite{Zou04} can be formulated in a variety of ways, and unlike PCA these method do not yield similar set of solutions. The formulation used by us involves treating Sparse PCA as a regression problem involving PCA, and using elastic net penalty to impose sparsity. In, this method PCA is computed by performing SVD on data matrix $\textbf{X}$ as:
 \begin{equation}\label{pca}
     X=UDV^{T}
 \end{equation}
 Here, $Z_{l}=U_{l}D_{ll}$ represents the principal components (PC) of each observation, and $V_{l}$ represents the loadings of principle component. $Z_{l}$ can subsequently be obtained by projecting input data $X$ on vector $V_{l}$ (i.e $XV_{l}=Z_{l}$) using equation(\ref{pca}). Based on this observation PCA can be treated as a regression problem, in which $Z_l$ represents the response vector and $V_{l}$ represents the regression coefficient.
 \begin{equation}
\hat{\beta}=\underset{\beta}{\textit{arg\:min}}||Z_{l}-\textbf{X}\beta||^{2}+\lambda||\beta||^{2}+\lambda_{1}||\beta||_{1}
\end{equation}
Here, $\hat{\textit{V}_{l}}=\frac{\hat{\beta}}{||\hat{\beta}||}$ is an sparse approximation of $V_{l}$ the original loading vector. These loading vector are combined to form a combined sparse representation of vector $V$ having $\textbf{L}$ component.
\begin{equation}
    \hat{V}=[\hat{V}_1 \;\hat{V}_2\; ... \;\hat{V}_l\; ... \;\hat{V}_L]
\end{equation}

Using the concept of Sparse PCA, the input HOG feature vector $X$ is projected on the approximated Sparse components as follows: 
\begin{equation}
    \alpha=X\hat{V}
\end{equation}
\begin{equation}
\alpha = 
\begin{pmatrix}
\alpha_{1,1}^{1} & \alpha_{1,2}^{1} & \cdots & \alpha_{1,L}^{1} \\
\alpha_{2,1}^{1} & \alpha_{2,2}^{1} & \cdots & \alpha_{2,L}^{1} \\
\vdots  & \vdots  & \vdots & \vdots  \\
\alpha_{N,1}^{1} & \alpha_{N,2}^{1} & \cdots & \alpha_{N,L}^{1} \\

\alpha_{1,1}^{2} & \alpha_{1,2}^{2} & \cdots & \alpha_{1,L}^{2} \\
\alpha_{2,2}^{2} & \alpha_{2,2}^{2} & \cdots & \alpha_{2,L}^{2} \\
\vdots  & \vdots  & \vdots & \vdots  \\
\alpha_{N,1}^{2} & \alpha_{N,2}^{2} & \cdots & \alpha_{N,L}^{2} \\

\alpha_{1,1}^{W} & \alpha_{1,2}^{W} & \cdots & \alpha_{1,L}^{W} \\
\alpha_{2,2}^{W} & \alpha_{2,2}^{W} & \cdots & \alpha_{2,L}^{W} \\
\vdots  & \vdots  & \vdots & \vdots  \\
\alpha_{N,1}^{W} & \alpha_{N,2}^{W} & \cdots & \alpha_{N,L}^{W} 
\end{pmatrix}
\end{equation}

Here, $\alpha_{j,k}^{i}$ represents the value of coefficient corresponding to the  $j^{th}$ Sparse component relative to the $k^{th}$ fragment of $i^{th}$ writer. Here, $N$ represents the  total number of fragments contributed by each writer for constructing sparse representation.\par

Coefficients generated by $W$ writers each contributing $N$ number of fragments (M = W × N) is used to construct a histogram for each of the individual components of the sparse vector using equation (\ref{hist}).
\begin{equation}\label{hist}
    H^{i}_{jb}=
\begin{cases}
    H^{i}_{jb}+1,& \text{if } h_{b}\leq\alpha_{j,k}^{i}<h_{b+1} \\
    H^{i}_{jb},              & \text{otherwise}
\end{cases}
\end{equation}

 Here, $b$ represents the value of quantize histogram bins, and $ H^{i}_{jb}$ signifies the number of nonzero entries for the $j$ sparse component of a writer $i$. Following the histogram generation each of histogram is normalized in the range between 0 and 1 as follows:
\begin{equation}
    p^{i}_{jb}=\frac{H^{i}_{jb}}{\sum^{B}_{b=1}H^{i}_{jb}}
\end{equation}

\subsection{Computation of filter saliency based on Entropy}
The normalized histogram $p^{i}_{jb}$ is used to compute entropy of the  principle components for each enrolled writer as:
\begin{equation}
    E^{i}_{j}=\sum_{b=1}^{B} -p^{i}_{jb}\:\mathrm{log_2}\:(p^{i}_{jb}) \quad 1\leq j\leq L
\end{equation}

Here, $E^{i}_{j}$ is the entropy value of the $j^{th}$ principle component corresponding to $N$ samples of $i^{th}$ enrolled writer in $W$. This entropy can be represented with the help of following entropy matrix:
\begin{equation}\label{Entr}
E = 
\begin{pmatrix}
E^{1}_{1} & E^{1}_{2} & \cdots &E^{1}_{L} \\
E^{2}_{1} & E^{2}_{2} & \cdots &E^{2}_{L} \\
\vdots  & \vdots  & \ddots & \vdots  \\
E^{W}_{1} & E^{W}_{2} & \cdots &E^{W}_{L}
\end{pmatrix}
\end{equation}
In this matrix the entries along the row specifies the entropy value relative to each sparse principle component of the enrolled writer $W$, and the entries along the column specifies the entropy for a particular sparse component relative to each of the enrolled writer. These entropy values gives us the information about the probability distribution of the principal components corresponding to each writers. These individual entropy values are summed up across all writers to give an overall entropy value for a particular convolution filter as follows:
\begin{equation}
    \phi_{f}=\frac{\sum_{w=1}^{W}\sum_{l=1}^{L}E_{j}^{i}}{W*L}
\end{equation}
Here, $\phi_{f}$ is the overall value of entropy for the $f^{th}$ filter of a convolution layer. This overall entropy value signifies the amount of information contained in a particular filter of a convolution layer. A low entropy value signifies the presence of information content across some of the principal components of the filter, thus making other components insignificant leading to a low entropy value. Such a filter should be assigned a low saliency value. On the contrary, a higher entropy value signifies, that the informative content is evenly distributed across a larger number of principle components thus corresponding filter needs to be assigned a higher saliency value. Based on these observations we assign saliency value to each of the filters using their entropy value as:
\begin{equation}\label{post}
    w_f=\frac{\phi_f}{\sum_f\phi_f}
\end{equation}
\par
\begin{figure*} 
\centering
\includegraphics[width=\textwidth]{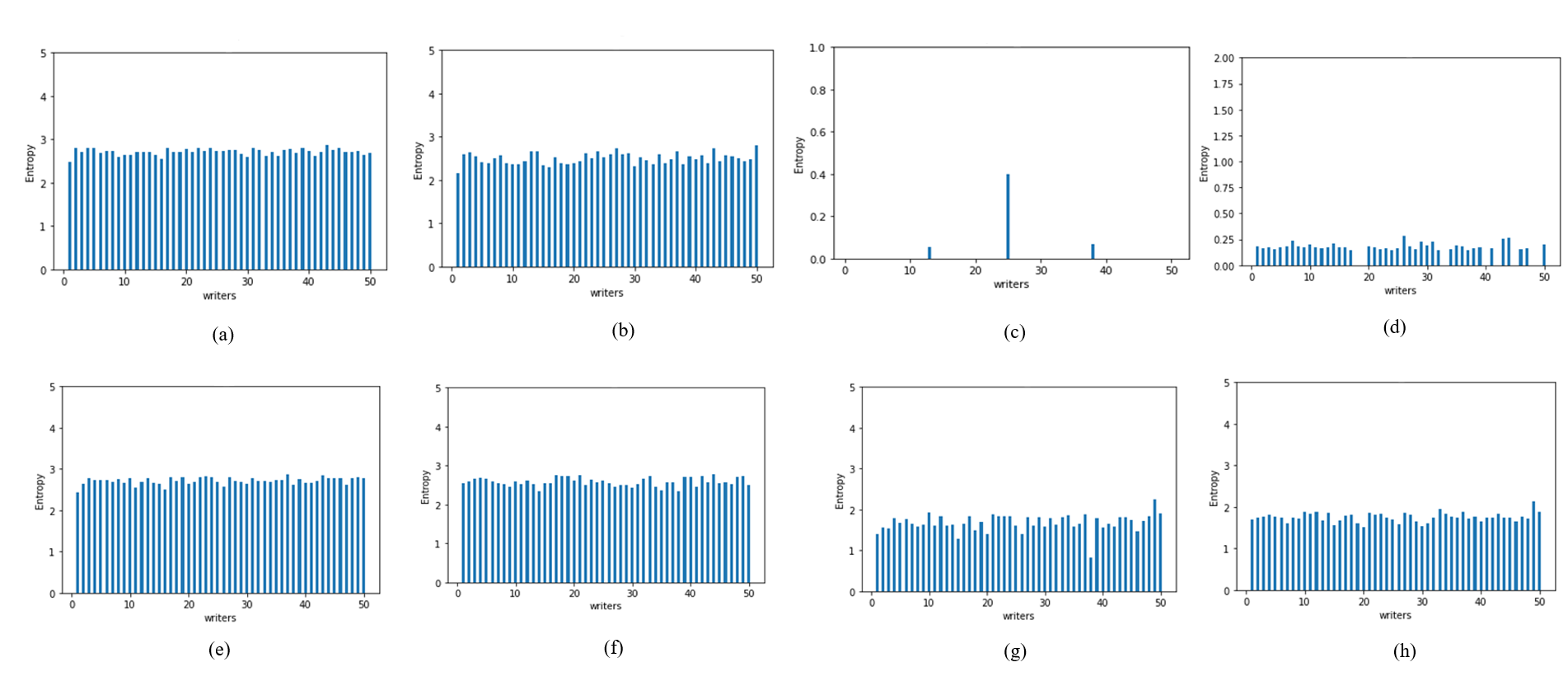}
\caption{(a) and (b) are histogram corresponding to two filters of conv1 having highest and second highest saliency values, (c) and (d) are histogram corresponding to two filters of conv1 having minimum and second minimum saliency values. Similarly (e) and (f) are histogram corresponding to two filters of conv2 having highest and second highest saliency values, (g) and (h) are histogram corresponding to filters of conv2 having minimum and second minimum saliency values.}
 \label{saliency}
\end{figure*}

As, a visual interpretation of the above explanation, we generate a histogram in Fig.\ref{saliency} corresponding to some of the filters of conv1 and conv2 layers in our convolution network. These histograms correspond to the filters having maximum and minimum saliency values. For constructing the histogram we have used 10 word samples from each of 50 writers of the IAM database\cite{IAM}. These value are arrived at by adding the entries along the rows in equation \ref{Entr}.

\section{writer feature generation and classification}\label{sec_7}
Given a handwritten document of a writer containing $N_{w}$ number of word samples, generating $w_{T}$ number of SIFT keypoints, which upon passing through a convolution network generates a set of feature map relative to the number of filters included in that layer as discussed in the earlier sections. These, feature map output are then the processed further to obtained an overall HoG representation for each fragment of a writer sample. \par
For obtaining an overall HOG feature representation form these feature map, we combine them using different types of pooling strategy. These are discussed below:
\begin{itemize}
    \item Average pooling: In this pooling methodology equal weightage is assigned to all the feature map of a convolution layer. These are then combined together to obtain an overall feature map.
    \begin{equation}\label{average}
        S^{j}_{p}=\sum_{f=1}^{F} (\frac{1}{F} \times z_{fp}^{j})
    \end{equation}
    Here, $z_{fp}^{j}$ represents the feature map corresponding to the $f^{th}$ filter of the convolution layer, for $j^{th}$ writer fragment, and $S^{j}_{p}$ represents the combined feature map. $p$ signifies the writer fragment. 
    \item Pre-saliency pooling: In this pooling strategy, a saliency value is assigned to the feature map of a convolution layer based on their entropy value as discussed in section \ref{sec_6}.
    \begin{equation}\label{pre}
         \tilde{S}^{j}_{p}=\sum_{f=1}^{F} (w_f \times z_{fp}^{j})
    \end{equation}
    Here, $w_f$ is the entropy-based saliency value assigned to each filter $f$ of a convolution layer, and $\tilde{S}^{j}_{p}$ represents the combined feature map output for the $p^{th}$ fragment of a writer. 
\end{itemize}
After, obtaining the combined feature map using the above mentioned pooling strategies, these are then transformed into one dimensional feature vector using modified HoG representation as discussed in section \ref{sec_5}. \par
 In addition to the above two pooling strategy. In our work, we also employed a third typed of pooling Strategy. In this pooling technique, instead of combining the feature map for generating HOG feature representation as done in above two pooling strategies, HoG feature is extracted individually for each feature map of a convolution layer, which are subsequently combined using corresponding saliency values of each filter in a convolution layer. Thus, in this technique pooling operation is performed at the later stages. This post-pooling based saliency technique is mathematically represented as:
 \begin{equation}\label{post}
     \hat{S}^{j}_{p}=\frac{\sum_{f=1}^{F} (w_f \times FV_{fp}^{j})}{||\sum_{f=1}^{F} (w_f \times FV_{fp}^{j})||_2}
 \end{equation}
Here, $FV_{fp}^{j}$ represents extracted HoG feature corresponding to the $f^{th}$ feature map output of the $j^{th}$ writer fragment, and $w_{f}$ represents the saliency value assigned to the $f^{th}$ filter output of the convolution layer. The effect on classification accuracy by incorporating these pooling strategy is discussed in section \ref{sec_9}. \par

Following the extraction of HOG feature vector, a one vs all SVM using radial basis function (RBF) \cite{Burges:1998} is used to train writing samples. The optimal values of RBF parameters $C$ and $\gamma$ are obtained by grid search. \par

During the testing phase, HOG feature vector generated from the SIFT keypoint fragments of a writer sample is passed is passed through a set of SVM classifiers to obtain a set of classification score for each of the word fragments. An, SVM classifier assigns a positive or negative value to input samples based on its proximity to the hyperplane separating postitve and negative training samples. These, SVM values are bounded in the range between $[0-1]$ by passing the  SVM classifier value through a sigmoidal function. Given, a word image containing $\hat{N}$ fragments, the normalised SVM score generated for each of these fragments is average pooled across each enrolled writer trained SVM classifier to obtain an overall response for the segmented word fragments.
\begin{equation}
    P(s)=\frac{1}{\hat{N}}\sum p_n
\end{equation}
Here, ($\hat{N}$) represents the total number of SIFT fragments generated form an input word image, $p_n$ is the normalized SVM score of the $n^{th}$ fragment for a writer. The final prediction  for a word sample ($s$) is made based on the label ($l$) for which the classifier reports the highest confidence score as:
\begin{equation}
 \hat{y}=\underset{l\in \{1,...,W\}}{argmax}P_{l}(s)
\end{equation}

\section{Dataset description}\label{sec_8}
The proposed method is evaluated on two datsets: IAM \cite{IAM}and CVL \cite{CVL}.\par

IAM \cite{IAM} dataset is the most widely used English language dataset for the purpose of writer identification. It contains 1593 images of English handwriting documents collected from a set of 657 writers. Each writer has contributed a variable number of handwritten document. Out of 657 writers, 301 have contributed two or more handwritten documents, while the rest have contributed only a single handwritten document. We modified the IAM dataset as described in \cite{Wu:2014} by randomly selecting two documents for writers who have contributed two or more document samples in the database. On the other hand for a writer contributing only one document, it is divided into half one half of which is used for training and the other half for testing. As the bounding boxes for the word image have been provided in the dataset, we collect the word image from the generated training and testing set for our experimentation.\par

CVL \cite{CVL} dataset contains handwritten text documents from 310 writers of which 27 writers have contributed 7 text documents, while the rest have contributed 5 text documents. Each writer contributed one text document in German and the rest in English language. In our experimentation, we used only the text document written in English. We follow the same methodology as is done in \cite{He:2020} for dividing the dataset into training and testing sets. Similar to the IAM dataset segmented words are made available in this dataset. \par

Table \ref{datset} gives a detailed overview of all the datasets used in our experiments.
\begin{table}[t]
\centering
\caption{Overview of the datasets used in experiments}
\label{datset}
\begin{tabular}{@{}ccc@{}}
\toprule
Dataset   & Number of writers & Language \\ \midrule
IAM       & 657               & English  \\
CVL       & 310               & English  \\ \bottomrule
\end{tabular}
\end{table}

\section{Experiment and Discussion}\label{sec_9}
In this section we carry out a set of experiments of our proposed algorithm on two benchmark datasets. These set of experiments helps us in choosing a proper set of parameters to be used for increasing the efficacy of our proposed writer identification system. 
\subsection{Implementation details}
The proposed Convolution network is built on the TensorFlow framework. Adam optimizer is used for optimizing the network, with a weight decay factor of $0.1$ every ten epoch. The model is trained for 50 epochs. Word fragments are normalized (between$[0-1]$) before feeding to the convolution network for feature extraction.\par
In order to extract the HOG feature vector from SIFT keypoint as discussed in section \ref{sec_5}, we select the value of $(m,n)$ to be $(4,4)$ which divides the cov1 and conv2 feature map into 16 sub-blocks. These 16 sub-blocks are grouped into $t=4$ quadrants, with each quadrant containing a combination of $b=4$ sub-blocks. Since, the size of the feature map in a convolution network decreases with increasing depth of the network, in order to maintain the spacial consistency of the feature map value of $m$ and $n$ is adjusted accordingly. For extracting the HOG features from conv3 feature map the value of $(m,n)$ is set to $(2,2)$, as the size of the conv3 layer is half the size of the conv2 layer. Similarly, the value of $t$ and $b$ is also adjusted accordingly.

\subsection{Performance with varying Bins B and convolution layers}
In this subsection, we evaluate the effect of bin size ($B$) on the overall performance of our writer identification system on approximately one third dataset (100 for CVL and 200 for IAM). We vary the value of bin size ($B$) from 6 to 12 in steps of two. Table \ref{res1} tabulates the result of our strategy for different databases. We observe that the best average identification rate for the IAM database corresponding to conv1 and conv2 layers are 91.15 and 90.25 respectively which is achieved for bin size 10. Similarly, for the CVL database, we achieve the best average accuracy of 82.97 and 80.95 corresponding to bin size 10. As, we move into con3 layer feature map, the identification rate decreases with respect to conv1 and conv2 layers.\par 

Based on the above tabulated results, it can be inferred that initial layers of the convolution neural network (namely conv1 and conv2) capture most of the information that is helpful in differentiating one set of writers from another. As we move deeper into the convolution network the resolution of the image decreases, thus, the underlying information becomes more abstract and less visual interpretative resulting in loss of discriminability among writers(as shown using conv3 result). In the subsequent set of experiments, we will mainly lay emphasis on conv1 and conv2 layers corresponding to the optimal bin size (as reported in Table \ref{res1}).

\begin{table}[t]
\centering
\caption{Comparison of average identification rates (in \% for one third data set samples) at word level for proposed algorithm with different bin size. The best identification rate is marked in bold}
\label{res1}
 \resizebox{.8\textwidth}{!}{%
\begin{tabular}{cccccc}
\hline
\multirow{2}{*}{Convolution layer} & \multirow{2}{*}{Bin size} & \multicolumn{2}{c}{IAM} & \multicolumn{2}{c}{CVL} \\ \cline{3-6} 
                                   &                           & Top1       & Top5       & Top1       & Top5       \\ \hline
\multirow{4}{*}{Conv1}             & 6                         & 89.8       & 95.45      & 81         & 92.45      \\
                                   & 8                         & 90.15      & 95.85      & 82         & 93.68      \\
                                   & 10                        & \textbf{91.15}      & 97.15      & \textbf{82.97}      & 94.59      \\
                                   & 12                        & 89.85      & 95.65      & 81.21      & 93.25      \\ \hline
\multirow{4}{*}{Conv2}             & 6                         & 89.4       & 95.5       & 79         & 92.35      \\
                                   & 8                         & 89.8       & 95.3       & 79.95      & 93.25      \\
                                   & 10                        & \textbf{90.25}      & 96.75      &\textbf{80.95}        & 93.58      \\
                                   & 12                        & 88.55      & 95.15       & 80.80      & 93.38      \\ \hline
\multirow{4}{*}{Conv3}             & 6                         & 70           &88.15            & 47      & 77.85           \\
                                   & 8                         & 73.35            & 89.81           &57.8            &83.25            \\
                                   & 10                        & 79.97            & 91.77           & 64.45           &87.9            \\
                                   & 12                        &81.88            &92.07            & 67.15           &88.55            \\ \hline
\end{tabular}}
\end{table}

\subsection{Influence of weighting strategy}
In this experiment, we have consider the effect of incorporating different weighting strategy on the overall accuracy of the proposed writer identification system. These weighting strategies involve: mean-pooling (\ref{average}), pre-pooling (\ref{pre}) and post-pooling (\ref{post}). A detailed discussion on these pooling strategies is provided in section \ref{sec_7}. \par

For analyzing the effect of accuracy using various pooling methodology, we randomly select a set of 10 word samples per enrolled writer in the database. The result obtained by analysing various pooling strategy is tabulated is table\ref{weighting}. Based on table \ref{weighting} we observe, that the performance of mean pooling methodology is lowest when compared to other two pooling strategy. This, is due to the fact that mean pooling assigns equal importance to all the filters of a convolution layers, due to which additional information contained in some of the convolution filter is not utilized properly. On, the other hand other two pooling strategy (pre and post) uses saliency based method that estimates in advance the relative importance of each convolution filter output. Thus, providing extra information related to the filter output, which is helpful in increasing the overall accuracy of the system. Experimental result also reveal that among the two pooling strategy (pre and post), post-pooling performs better than pre-pooling. This, can be attributed to the fact that various filter output of a convolution layer are complementary to each other, as such using saliency values beforehand sometimes leads to masking of some important features in the combined feature map resulting in a loss of some information. On, the other hand in post-pooling strategy weights are assigned to the extracted HOG feature in a convolution layer separately, as such some hidden information can be captured effectively leading to increase in model accuracy.   

\begin{table}[t]
\centering
\caption{Effect of accuracy achieved (in \%) on IAM and CVL dataset using various weighting strategies }
\label{weighting}
 \resizebox{\textwidth}{!}{%
\begin{tabular}{cccc|ccc}
\hline
& \multicolumn{3}{c|}{Conv1} & \multicolumn{3}{c}{Conv2}     \\ \hline
Database  & Average weighting & pre weighting & post weighting & Average weighting & pre weighting & post weighting \\ \hline
IAM       & 84.77                   & 85.02  &86.07                &84.42             & 84.62          & 85.81                \\
CVL       & 77.53                   &78.05       & 78.24                & 75.29            & 75.54           & 77.79                \\ \hline
              
\end{tabular}}
\end{table}

\subsection{Score Fusion}
Let $I$ be an input word image generating $w_T$ number of HoG features corresponding to conv1 and conv2 layers respectively. These HOG feature when passed through a trained SVM classifier trained on  conv1 and conv2 layer HOG features generates $P_1^{I}(s)$ and $P_2^{I}(s)$ score respectively. These scores are then fused together to generate a final classification score $D^{I}$ as:
\begin{equation}
    D^{I}(s)=\alpha P_1^{I}(s)+(1-\alpha)P_2^{I}(s)
\end{equation}
Where, $0\leq \alpha\leq 1$ is the weight. The component $\alpha$ assigns a weight to the individual conv1 and conv2 classifiers by taking into account the effectiveness of the individual classifier in classifying a writer. The parameter $\alpha$ is database dependent and is determined based on validation data of one third enrolled writer in the database. \par
Once, the parameter $\alpha$ is selected the final prediction score for each writer is calculated individually, following which the prediction for candidate writer $j^{*}$ is made by the system as follows:
\begin{equation}
    j^{*}=\underset{j}{arg\:max}\;D^{I}(s,j)
\end{equation}

\subsection{Performance of Writer Identification with varying number of words Images}
\begin{figure*}[h] 
\centering
\includegraphics[width=\textwidth]{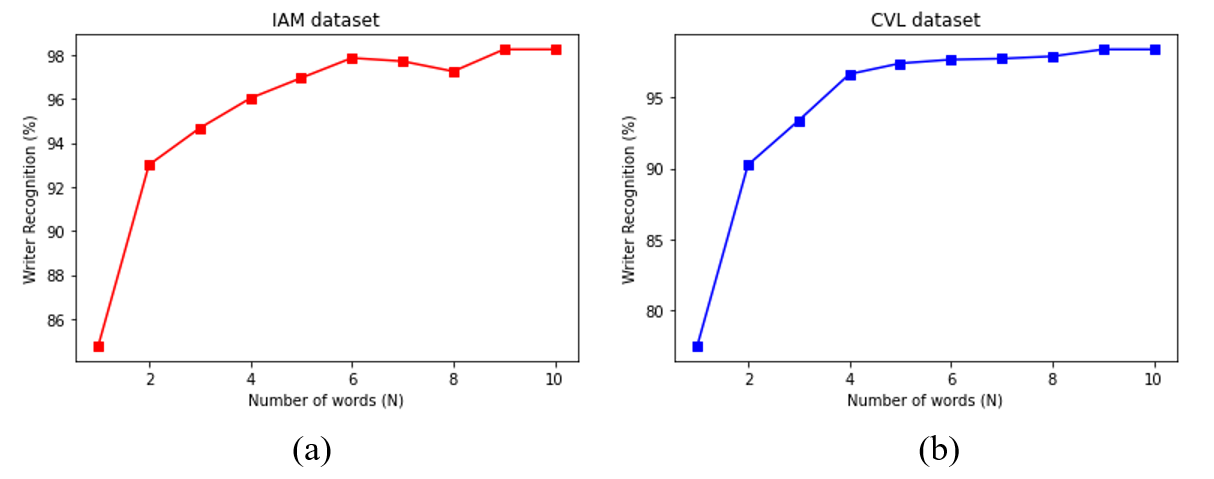}
\caption{Performance (Top1) of writer identification using different numbers of words on the IAM and CVL data set. }
\label{var_wrd}
\end{figure*}

In order to show the robustness of our proposed system with respect to the amount of handwritten word available, we calculate the TOP-1 performance of our system with respect to different number of handwritten words across two databases. In this experiment we randomly select $N$ number of word images for each enrolled writer. SIFT fragment extracted from these word samples is passed through HOG classifier corresponding to conv1 and conv2 layers following which these scores are combined to obtain overall accuracy score. This procedure is repeated 10 times and the average result for different $N$ values is shown with the help of fig.\ref{var_wrd}.\par

From fig.\ref{var_wrd}, it can be inferred that the performance of proposed writer identification system increases significantly as the number of available word sample of a writer increases form one to two, and it stabilizes as the number of available word sample reaches four or more words. This result shows the efficacy of our proposed algorithm in recognising writer in situation where limited amount of writer data is available for testing. 

\subsection{Performance comparison of proposed system with different features on word Image}
In this section we evaluate the performance of our proposed system with traditional handcrafted features based system on word image. Table \ref{result1} summarizes the result of different writer identification system considering word image as input. From the table, it can be observe that the performance of traditional handcrafted features based system on input word image is low. This is due to the fact that these algorithm make writer prediction based on statistical information collected from the input image. In order to generate stable representation based on statistical information, these algorithms require certain minimum amount of text samples. In the case of word image the statistical information captured is insufficient to generate a stable representation of the input text sample. On the contrary due to the presence of multiple filter in each convolution layer of a convolution network, it can capture a diverse quantity of information from the training data, thus increasing the overall performance of the system. In addition to this, it can also be deduced form table \ref{result1},that a convolution network train on writer-independent auxiliary dataset is able to capture general writer characteristics more effectively compared to a network trained on a writer dependent dataset.

\begin{table}[t]
\centering
\caption{Comparison of Writer Identification performance (in \% ) on full database for word image samples}
\label{result1}
 \resizebox{\textwidth}{!}{%
\begin{tabular}{ccccccc}
\hline
\multirow{2}{*}{Method}                  & \multicolumn{2}{c}{IAM}                     & \multicolumn{2}{c}{CVL}                                 \\ \cline{2-5} 
                                         & Top1                 & Top5                                  & Top1                 & Top5                                \\ \hline
Hinge\cite{Bulacu:2007}                                    & 13.8                 & 28.3                 & 13.6                 & 29.7                                                  \\
Quill\cite{Brink:2012}                                    & 23.8                 & 44.0                 & 23.8                 & 46.7                                                  \\
CoHinge\cite{Sheng:2017}                                 & 19.4                 & 34.1                 & 18.2                 & 34.2                                                  \\
QuadHinge\cite{Sheng:2017}                                & 20.9                 & 37.4                 & 17.8                 & 35.5                                                   \\
COLD\cite{sheng:2017C}                                    & 12.3                 & 28.3                 & 12.4                 & 29.0                                                  \\
Chain Code Pairs\cite{Siddiqi:2010}                         & 12.4                 & 27.1                 & 13.5                 & 30.3                                                   \\
Chain Code Triplets\cite{Siddiqi:2010}                       & 16.9                 & 33.0                 & 17.2                 & 35.4                                                   \\
WordImgNet\cite{He:2020}                                & 52.4                 & 70.9                 & 62.5                 & 82.0                                                   \\
FragNet-64\cite{He:2020}                               & 72.2                 & 88.0                 & 79.2                 & 93.3                                                  \\
Veritcal GR-RNN(FGRR)\cite{He:2021}                               & 83.3                 & 94.0                 & 83.5                 & 94.6                                                  \\
Horizontal GR-RNN(FGRR)\cite{He:2021}                                 & 82.4                 & 93.8                 &82.9                 & 94.6                                                  \\\hline
\multicolumn{1}{l}{Proposed Methodology with average pooling (conv1+conv2)} & \multicolumn{1}{l}{85.87} & \multicolumn{1}{l}{92.30} & \multicolumn{1}{l}{79.70} & \multicolumn{1}{l}{91.94}  \\
\multicolumn{1}{l}{Proposed Methodology with pre weighting (conv1+conv2)} & \multicolumn{1}{l}{86.16} & \multicolumn{1}{l}{92.77} & \multicolumn{1}{l}{80.16} & \multicolumn{1}{l}{92.14} \\
\multicolumn{1}{l}{Proposed Methodology with post weighting (conv1+conv2)} & \multicolumn{1}{l}{86.68} & \multicolumn{1}{l}{92.96} & \multicolumn{1}{l}{80.57} & \multicolumn{1}{l}{92.35}  \\ \hline
\end{tabular}}
\end{table}

\subsection{Performance comparison of proposed system with other methods on page Image}
In this section we evaluate the performance of our proposed algorithm on page images. For evaluating the writer performance on page images, we calculate the individual SVM classification score of each word image contained in the page of writer sample, following which these scores are aggregated to arrive at an overall classification score. This process is represented mathematical using below equation.  
\begin{equation}
    P_{page}=\frac{1}{N_{w}}\sum_{t\in page}^{N_{w}}P(t)
\end{equation}
where $P_{page}$ is the overall writer classification score generated based on words samples present in a $page$ image. $P(t)$ is the individual classification score of each word $t$ contained in a $page$ and $N_{w}$ is the total number of word in the $page$ image. Table \ref{sota_IAM} and \ref{sota_CVL} shows the overall performance of our algorithm  on page level data when compared with other state of art algorithms for IAM and CVL datasets.\par

\begin{figure}[h] 
\centering
\includegraphics[width=\textwidth]{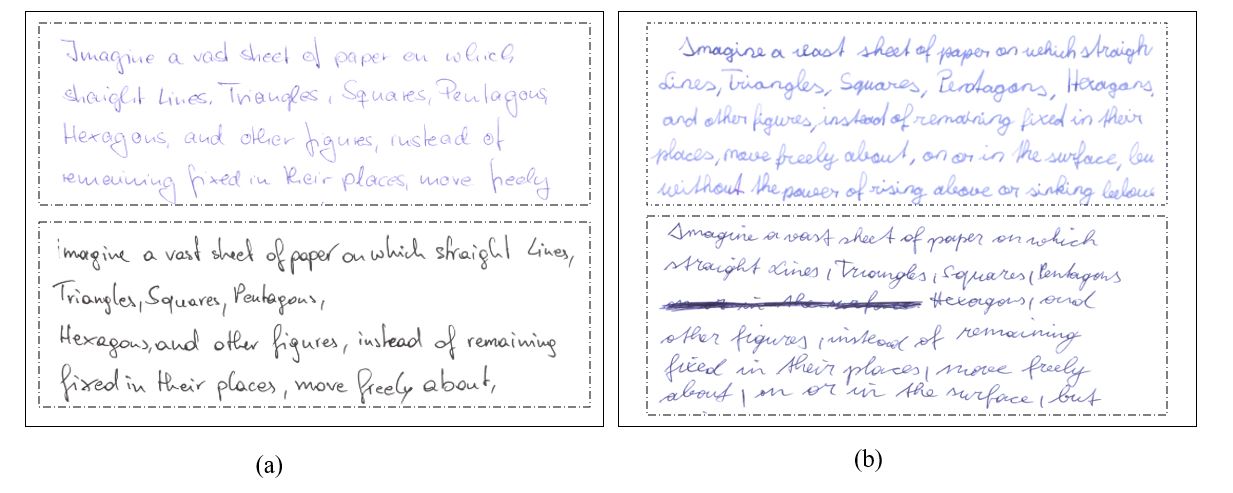}
\caption{Example of false acceptance. Samples included in (a) and (b) are from different writers, but is identified as same writer by the proposed algorithm.}
\label{false_acc}
\end{figure}

Based on table \ref{sota_IAM} and \ref{sota_CVL} it can be observed that the performance of the proposed writer identification algorithm on page image is better than word image due to the availability of large number of word samples contained in a page. Our word based writer identification algorithm achieves a higher identification accuracy when compared with other word based writer identification algorithm (such as WordImagenet, Fragnet\cite{He:2020}, GR-RNN \cite{He:2021}, and \cite{Nguyen:2018}) for page level data on IAM datset. For CVL\cite{CVL} dataset our proposed algorithm achieves top1 accuracy of 99.0\%, which is comparable to other state of art algorithm for page level data. The low classification accuracy of our proposed system on page level data for CVL dataset as compared to other algorithm is due to the fact that the written content in case of CVL datase is same for every writer due to which, the model in some of the case is unable to differentiate between writers having a similar style (as our proposed model is not trained to capture writer specific features), thus leading to false writer identification in some of the cases, as depicted with the help of figure \ref{false_acc}. This result further affirms the inference made in the previous discussion regarding the robustness of writer independent model in learning general writer characteristics.

\begin{table}[t]
\centering
\caption{ Comparison of State of art methods on IAM database}
\label{sota_IAM}
 \resizebox{\textwidth}{!}{%
\begin{tabular}{ccc}
\hline
Reference                                  & Number of Writers & Top 1 accuracy \\ \hline
Siddiqui and Vincent\cite{Siddiqi:2010}                      & 650               & 91.0           \\
He and Schomaker\cite{Sheng:2017}                           & 650               & 93.2           \\
Khalifa et.al.\cite{Khalifa:2015}                             & 650               & 92.0           \\
Hadjadji and Chibani\cite{Hadjadji:2018}                       & 657               & 94.5           \\
Wu et.al.\cite{Wu:2014}                                 & 657               & 98.5           \\
Khan et.al.\cite{Khan:2018}                               & 650               & 97.8           \\
Nguyen et.al.\cite{Nguyen:2018}                              & 650               & 93.1           \\
WordImagenet                               & 657               & 95.8           \\
FragNet-64\cite{He:2020}                                & 657               & 96.3           \\
GR-RNN\cite{He:2021}                                   & 657               & 96.4           \\
Proposed Methodology with Post weighting   & 657               & 98.17               \\ \hline
\end{tabular}}
\end{table}

\begin{table}[t]
\centering
\caption{Comparison of state of art method on CVL database}
\label{sota_CVL}
 \resizebox{\textwidth}{!}{%
\begin{tabular}{ccc}
\hline
Reference                                  & Number of Writers & Top 1 accuracy \\ \hline
Fiel and sablating\cite{Fiel:2015}                        & 309               & 97.8           \\
Tang and Wu\cite{Tang:2016}                               & 310               & 99.7           \\
Christlein et.al\cite{Christlein:2017}                           & 310               & 99.2           \\
Khan et.al.\cite{Khan:2018}                                & 310               & 99.0           \\
WordImagenet                               & 310               & 98.8           \\
FragNet-64\cite{He:2020}                                 & 310               & 99.1           \\
GR-RNN\cite{He:2021}                                     & 310               & 99.3           \\
Proposed Methodology with Post weighting   & 309               & 99.0              \\ \hline
\end{tabular}}
\end{table}

\section{Conclusion}\label{sec_10}
In this work we have proposed an offline test-independent writer identification system based on word image. Our algorithm uses SIFT keypoint to detect various type of features from a word image, these feature vary form word to allograph. Such type of diverse features helps in making our system more robust as compared to a system trained using word image as whole. The major contribution of our proposed work can be listed as follows: (i) Use of a convolution network trained on a auxiliary dataset instead of an end to end trained network on writer dependent data. Thus, making the algorithm better suited to extract writer features without the need of retraining the network. (ii) Use of modified HOG feature representation to capture spatial relation between convolution later output of various SIFT keypoint corresponding to different writers. (iii) proposing an entropy based weighting strategy to assign importance to each filter of a convolution layer. \par 
One of the limitation of our proposed algorithm is that it requires a good quality of segmented word image, which is a very challenging task to achieve on a highly cursive written document. In our future works we would like to address this shortcoming and work on an algorithm which could perform writer identification on any handwritten document.

\bibliographystyle{elsarticle-num-names} 
\bibliography{cas-refs}

\begin{thebibliography}{50}
\expandafter\ifx\csname natexlab\endcsname\relax\def\natexlab#1{#1}\fi
\providecommand{\url}[1]{\texttt{#1}}
\providecommand{\href}[2]{#2}
\providecommand{\path}[1]{#1}
\providecommand{\DOIprefix}{doi:}
\providecommand{\ArXivprefix}{arXiv:}
\providecommand{\URLprefix}{URL: }
\providecommand{\Pubmedprefix}{pmid:}
\providecommand{\doi}[1]{\href{http://dx.doi.org/#1}{\path{#1}}}
\providecommand{\Pubmed}[1]{\href{pmid:#1}{\path{#1}}}
\providecommand{\bibinfo}[2]{#2}
\ifx\xfnm\relax \def\xfnm[#1]{\unskip,\space#1}\fi
\bibitem[{Xu et~al.(2016)Xu, Fei, and Jie}]{Xu:2018}
\bibinfo{author}{Y.~Xu}, \bibinfo{author}{L.~Fei}, \bibinfo{author}{W.~Jie},
\newblock \bibinfo{title}{Discriminative and robust competitive code for
  palmprint recognition},
\newblock \bibinfo{journal}{IEEE Transactions on Systems, Man, and Cybernetics:
  Systems} \bibinfo{volume}{PP} (\bibinfo{year}{2016}) \bibinfo{pages}{1--10}.
\bibitem[{Chi et~al.(2018)Chi, Wang, and Meng}]{Chi:2018}
\bibinfo{author}{W.~Chi}, \bibinfo{author}{J.~Wang}, \bibinfo{author}{M.~Q.-H.
  Meng},
\newblock \bibinfo{title}{A gait recognition method for human following in
  service robots},
\newblock \bibinfo{journal}{IEEE Transactions on Systems, Man, and Cybernetics:
  Systems} \bibinfo{volume}{48} (\bibinfo{year}{2018})
  \bibinfo{pages}{1429--1440}.
\bibitem[{Fernandez-de Sevilla et~al.(2010)Fernandez-de Sevilla,
  Alonso-Fernandez, Fierrez, and Ortega-Garcia}]{Fernandez:2010}
\bibinfo{author}{R.~Fernandez-de Sevilla},
  \bibinfo{author}{F.~Alonso-Fernandez}, \bibinfo{author}{J.~Fierrez},
  \bibinfo{author}{J.~Ortega-Garcia},
\newblock \bibinfo{title}{Forensic writer identification using allographic
  features},
\newblock in: \bibinfo{booktitle}{2010 12th International Conference on
  Frontiers in Handwriting Recognition}, \bibinfo{year}{2010}, pp.
  \bibinfo{pages}{308--313}.
\bibitem[{Bulacu et~al.(2007)Bulacu, van Koert, Schomaker, and van~der
  Zant}]{Bulacu:2007}
\bibinfo{author}{M.~Bulacu}, \bibinfo{author}{R.~van Koert},
  \bibinfo{author}{L.~Schomaker}, \bibinfo{author}{T.~van~der Zant},
\newblock \bibinfo{title}{Layout analysis of handwritten historical documents
  for searching the archive of the cabinet of the dutch queen},
\newblock in: \bibinfo{booktitle}{Ninth International Conference on Document
  Analysis and Recognition (ICDAR 2007)}, volume~\bibinfo{volume}{1},
  \bibinfo{year}{2007}, pp. \bibinfo{pages}{357--361}.
\bibitem[{He et~al.(2014)He, Sammara, Burgers, and Schomaker}]{He:2014}
\bibinfo{author}{S.~He}, \bibinfo{author}{P.~Sammara},
  \bibinfo{author}{J.~Burgers}, \bibinfo{author}{L.~Schomaker},
\newblock \bibinfo{title}{Towards style-based dating of historical documents},
\newblock in: \bibinfo{booktitle}{2014 14th International Conference on
  Frontiers in Handwriting Recognition}, \bibinfo{year}{2014}, pp.
  \bibinfo{pages}{265--270}.
\bibitem[{Faundez-Zanuy et~al.(2020)Faundez-Zanuy, Fierrez, Ferrer, Diaz,
  Tolosana, and Plamondon}]{Faundez:2020}
\bibinfo{author}{M.~Faundez-Zanuy}, \bibinfo{author}{J.~Fierrez},
  \bibinfo{author}{M.~Ferrer}, \bibinfo{author}{M.~Diaz},
  \bibinfo{author}{R.~Tolosana}, \bibinfo{author}{R.~Plamondon},
\newblock \bibinfo{title}{Handwriting biometrics: Applications and future
  trends in e-security and e-health},
\newblock \bibinfo{journal}{Cognitive Computation} \bibinfo{volume}{12}
  (\bibinfo{year}{2020}).
\bibitem[{Sharma and Sundaram(2017)}]{Sharma:2017}
\bibinfo{author}{A.~Sharma}, \bibinfo{author}{S.~Sundaram},
\newblock \bibinfo{title}{A novel online signature verification system based on
  gmm features in a dtw framework},
\newblock \bibinfo{journal}{IEEE Transactions on Information Forensics and
  Security} \bibinfo{volume}{12} (\bibinfo{year}{2017})
  \bibinfo{pages}{705--718}.
\bibitem[{Xiong et~al.(2017)Xiong, Lu, and Wang}]{Xiong:2017}
\bibinfo{author}{Y.-J. Xiong}, \bibinfo{author}{Y.~Lu},
  \bibinfo{author}{P.~S.~P. Wang},
\newblock \bibinfo{title}{Off-line text-independent writer recognition: A
  survey},
\newblock \bibinfo{journal}{International Journal of Pattern Recognition and
  Artificial Intelligence} \bibinfo{volume}{31} (\bibinfo{year}{2017})
  \bibinfo{pages}{1756008}.
\bibitem[{Said et~al.(2000)Said, Tan, and Baker}]{SAID:2000}
\bibinfo{author}{H.~Said}, \bibinfo{author}{T.~Tan},
  \bibinfo{author}{K.~Baker},
\newblock \bibinfo{title}{Personal identification based on handwriting},
\newblock \bibinfo{journal}{Pattern Recognition} \bibinfo{volume}{33}
  (\bibinfo{year}{2000}) \bibinfo{pages}{149--160}.
\bibitem[{Bertolini et~al.(2013)Bertolini, Soares~de Oliveira, Justino, and
  Sabourin}]{Bertolini:2013}
\bibinfo{author}{D.~Bertolini}, \bibinfo{author}{L.~Soares~de Oliveira},
  \bibinfo{author}{E.~Justino}, \bibinfo{author}{R.~Sabourin},
\newblock \bibinfo{title}{Texture-based descriptors for writer identification
  and verification},
\newblock \bibinfo{journal}{Expert Systems with Applications}
  \bibinfo{volume}{40} (\bibinfo{year}{2013}) \bibinfo{pages}{2069–2080}.
\bibitem[{Heikkila and Ojansivu(2009)}]{Heikkila:2009}
\bibinfo{author}{J.~Heikkila}, \bibinfo{author}{V.~Ojansivu},
\newblock \bibinfo{title}{Methods for local phase quantization in
  blur-insensitive image analysis},
\newblock in: \bibinfo{booktitle}{2009 International Workshop on Local and
  Non-Local Approximation in Image Processing, LNLA 2009},
  \bibinfo{year}{2009}, pp. \bibinfo{pages}{104 -- 111}.
\bibitem[{Djeddi et~al.(2013)Djeddi, Siddiqi, Souici-Meslati, and
  Ennaji}]{Djeddi:2013}
\bibinfo{author}{C.~Djeddi}, \bibinfo{author}{I.~Siddiqi},
  \bibinfo{author}{L.~Souici-Meslati}, \bibinfo{author}{A.~Ennaji},
\newblock \bibinfo{title}{Text-independent writer recognition using
  multi-script handwritten texts},
\newblock \bibinfo{journal}{Pattern Recognition Letters} \bibinfo{volume}{34}
  (\bibinfo{year}{2013}).
\bibitem[{Bulacu et~al.(2003)Bulacu, Schomaker, and Vuurpijl}]{Bulacu:2003}
\bibinfo{author}{M.~Bulacu}, \bibinfo{author}{L.~Schomaker},
  \bibinfo{author}{L.~Vuurpijl},
\newblock \bibinfo{title}{Writer identification using edge-based directional
  features},
\newblock in: \bibinfo{booktitle}{Seventh International Conference on Document
  Analysis and Recognition, 2003. Proceedings.}, \bibinfo{year}{2003}, pp.
  \bibinfo{pages}{937--941}.
\bibitem[{Hannad et~al.(2016)Hannad, Siddiqi, and Elkettani}]{Hannad:2016}
\bibinfo{author}{Y.~Hannad}, \bibinfo{author}{I.~Siddiqi},
  \bibinfo{author}{Y.~Elkettani},
\newblock \bibinfo{title}{Writer identification using texture descriptors of
  handwritten fragments},
\newblock \bibinfo{journal}{Expert Systems with Applications}
  \bibinfo{volume}{47} (\bibinfo{year}{2016}) \bibinfo{pages}{14--22}.
\bibitem[{Wu et~al.(2014)Wu, Tang, and Bu}]{Wu:2014}
\bibinfo{author}{X.~Wu}, \bibinfo{author}{Y.~Tang}, \bibinfo{author}{W.~Bu},
\newblock \bibinfo{title}{Offline text-independent writer identification based
  on scale invariant feature transform},
\newblock \bibinfo{journal}{IEEE Transactions on Information Forensics and
  Security} \bibinfo{volume}{9} (\bibinfo{year}{2014})
  \bibinfo{pages}{526--536}.
\bibitem[{Khan et~al.(2018)Khan, Khelifi, Tahir, and Bouridane}]{Khan:2018}
\bibinfo{author}{F.~Khan}, \bibinfo{author}{F.~Khelifi},
  \bibinfo{author}{M.~Tahir}, \bibinfo{author}{A.~Bouridane},
\newblock \bibinfo{title}{Dissimilarity gaussian mixture models for efficient
  offline handwritten text-independent identification using sift and rootsift
  descriptors},
\newblock \bibinfo{journal}{IEEE Transactions on Information Forensics and
  Security} \bibinfo{volume}{PP} (\bibinfo{year}{2018}) \bibinfo{pages}{1--1}.
\bibitem[{Newell and Griffin(2014)}]{Newell:2014}
\bibinfo{author}{A.~J. Newell}, \bibinfo{author}{L.~D. Griffin},
\newblock \bibinfo{title}{Writer identification using oriented basic image
  features and the delta encoding},
\newblock \bibinfo{journal}{Pattern Recogn.} \bibinfo{volume}{47}
  (\bibinfo{year}{2014}) \bibinfo{pages}{2255–2265}. \URLprefix
  \url{https://doi.org/10.1016/j.patcog.2013.11.029}.
  \DOIprefix\doi{10.1016/j.patcog.2013.11.029}.
\bibitem[{Schomaker and Bulacu(2004)}]{Schomaker:2004}
\bibinfo{author}{L.~Schomaker}, \bibinfo{author}{M.~Bulacu},
\newblock \bibinfo{title}{Automatic writer identification using
  connected-component contours and edge-based features of uppercase western
  script},
\newblock \bibinfo{journal}{IEEE Transactions on Pattern Analysis and Machine
  Intelligence} \bibinfo{volume}{26} (\bibinfo{year}{2004})
  \bibinfo{pages}{787--798}.
\bibitem[{Siddiqi and Vincent(2010)}]{Siddiqi:2010}
\bibinfo{author}{I.~Siddiqi}, \bibinfo{author}{N.~Vincent},
\newblock \bibinfo{title}{Text independent writer recognition using redundant
  writing patterns with contour-based orientation and curvature features},
\newblock \bibinfo{journal}{Pattern Recognition} \bibinfo{volume}{43}
  (\bibinfo{year}{2010}) \bibinfo{pages}{3853--3865}.
\bibitem[{Schomaker et~al.(2015)Schomaker, Wiering, and Sheng}]{Schomaker:2015}
\bibinfo{author}{L.~Schomaker}, \bibinfo{author}{M.~Wiering},
  \bibinfo{author}{H.~Sheng},
\newblock \bibinfo{title}{Junction detection in handwritten documents and its
  application to writer identification},
\newblock \bibinfo{journal}{Pattern Recognition} \bibinfo{volume}{48}
  (\bibinfo{year}{2015}).
\bibitem[{Akram et~al.(2019)Akram, Djeddi, Gattal, Siddiqi, and
  Tahar}]{Akram:2019}
\bibinfo{author}{B.~Akram}, \bibinfo{author}{C.~Djeddi},
  \bibinfo{author}{A.~Gattal}, \bibinfo{author}{I.~Siddiqi},
  \bibinfo{author}{M.~Tahar},
\newblock \bibinfo{title}{Handwriting based writer recognition using implicit
  shape codebook},
\newblock \bibinfo{journal}{Forensic Science International}
  \bibinfo{volume}{301} (\bibinfo{year}{2019}).
\bibitem[{Fiel and Sablatnig(2015)}]{Fiel:2015}
\bibinfo{author}{S.~Fiel}, \bibinfo{author}{R.~Sablatnig},
\newblock \bibinfo{title}{Writer identification and retrieval using a
  convolutional neural network},
\newblock in: \bibinfo{booktitle}{International Conference on Computer Analysis
  of Images and Patterns}, \bibinfo{organization}{Springer},
  \bibinfo{year}{2015}, pp. \bibinfo{pages}{26--37}.
\bibitem[{Xing and Qiao(2016)}]{Xing:2016}
\bibinfo{author}{L.~Xing}, \bibinfo{author}{Y.~Qiao},
\newblock \bibinfo{title}{Deepwriter: A multi-stream deep cnn for
  text-independent writer identification},
\newblock in: \bibinfo{booktitle}{2016 15th International Conference on
  Frontiers in Handwriting Recognition (ICFHR)}, \bibinfo{year}{2016}, pp.
  \bibinfo{pages}{584--589}.
\bibitem[{Christlein et~al.(2017)Christlein, Gropp, Fiel, and
  Maier}]{christlein:2017(b)}
\bibinfo{author}{V.~Christlein}, \bibinfo{author}{M.~Gropp},
  \bibinfo{author}{S.~Fiel}, \bibinfo{author}{A.~Maier},
\newblock \bibinfo{title}{Unsupervised feature learning for writer
  identification and writer retrieval},
\newblock in: \bibinfo{booktitle}{2017 14th IAPR International Conference on
  Document Analysis and Recognition (ICDAR)}, volume~\bibinfo{volume}{1},
  \bibinfo{organization}{IEEE}, \bibinfo{year}{2017}, pp.
  \bibinfo{pages}{991--997}.
\bibitem[{Yang et~al.(2016)Yang, Jin, and Liu}]{Yang:2016}
\bibinfo{author}{W.~Yang}, \bibinfo{author}{L.~Jin}, \bibinfo{author}{M.~Liu},
\newblock \bibinfo{title}{Deepwriterid: An end-to-end online text-independent
  writer identification system},
\newblock \bibinfo{journal}{IEEE Intelligent Systems} \bibinfo{volume}{31}
  (\bibinfo{year}{2016}) \bibinfo{pages}{45--53}.
\bibitem[{Rehman et~al.(2019)Rehman, Naz, Razzak, and Hameed}]{Rehman:2019}
\bibinfo{author}{A.~Rehman}, \bibinfo{author}{S.~Naz}, \bibinfo{author}{M.~I.
  Razzak}, \bibinfo{author}{I.~A. Hameed},
\newblock \bibinfo{title}{Automatic visual features for writer identification:
  A deep learning approach},
\newblock \bibinfo{journal}{IEEE Access} \bibinfo{volume}{7}
  (\bibinfo{year}{2019}) \bibinfo{pages}{17149--17157}.
\bibitem[{Sulaiman et~al.(2019)Sulaiman, Omar, Nasrudin, and
  Arram}]{Sulaiman:2019}
\bibinfo{author}{A.~Sulaiman}, \bibinfo{author}{K.~Omar},
  \bibinfo{author}{M.~F. Nasrudin}, \bibinfo{author}{A.~Arram},
\newblock \bibinfo{title}{Length independent writer identification based on the
  fusion of deep and hand-crafted descriptors},
\newblock \bibinfo{journal}{IEEE Access} \bibinfo{volume}{7}
  (\bibinfo{year}{2019}) \bibinfo{pages}{91772--91784}.
\bibitem[{He and Schomaker(2019)}]{He:2019}
\bibinfo{author}{S.~He}, \bibinfo{author}{L.~Schomaker},
\newblock \bibinfo{title}{Deep adaptive learning for writer identification
  based on single handwritten word images},
\newblock \bibinfo{journal}{Pattern Recognition} \bibinfo{volume}{88}
  (\bibinfo{year}{2019}) \bibinfo{pages}{64--74}.
\bibitem[{He and Schomaker(2020)}]{He:2020}
\bibinfo{author}{S.~He}, \bibinfo{author}{L.~Schomaker},
\newblock \bibinfo{title}{Fragnet: Writer identification using deep fragment
  networks},
\newblock \bibinfo{journal}{IEEE Transactions on Information Forensics and
  Security} \bibinfo{volume}{15} (\bibinfo{year}{2020})
  \bibinfo{pages}{3013--3022}.
\bibitem[{He and Schomaker(2021)}]{He:2021}
\bibinfo{author}{S.~He}, \bibinfo{author}{L.~Schomaker},
\newblock \bibinfo{title}{Gr-rnn: Global-context residual recurrent neural
  networks for writer identification},
\newblock \bibinfo{journal}{Pattern Recognition} \bibinfo{volume}{117}
  (\bibinfo{year}{2021}) \bibinfo{pages}{107975}.
\bibitem[{Felzenszwalb et~al.(2010)Felzenszwalb, Girshick, McAllester, and
  Ramanan}]{Felzenszwalb:2009}
\bibinfo{author}{P.~F. Felzenszwalb}, \bibinfo{author}{R.~B. Girshick},
  \bibinfo{author}{D.~McAllester}, \bibinfo{author}{D.~Ramanan},
\newblock \bibinfo{title}{Object detection with discriminatively trained
  part-based models},
\newblock \bibinfo{journal}{IEEE Transactions on Pattern Analysis and Machine
  Intelligence} \bibinfo{volume}{32} (\bibinfo{year}{2010})
  \bibinfo{pages}{1627--1645}.
\bibitem[{Lowe(2004)}]{Lowe:2004}
\bibinfo{author}{D.~G. Lowe},
\newblock \bibinfo{title}{Distinctive image features from scale-invariant
  keypoints},
\newblock \bibinfo{journal}{Int. J. Comput. Vision} \bibinfo{volume}{60}
  (\bibinfo{year}{2004}) \bibinfo{pages}{91–110}.
  \DOIprefix\doi{10.1023/B:VISI.0000029664.99615.94}.
\bibitem[{Zhang et~al.(2003)Zhang, Srihari, and Lee}]{zhang:2003}
\bibinfo{author}{B.~Zhang}, \bibinfo{author}{S.~N. Srihari},
  \bibinfo{author}{S.~Lee},
\newblock \bibinfo{title}{Individuality of handwritten characters},
\newblock in: \bibinfo{booktitle}{Seventh International Conference on Document
  Analysis and Recognition, 2003. Proceedings.}, volume~\bibinfo{volume}{3},
  \bibinfo{organization}{IEEE Computer Society}, \bibinfo{year}{2003}, pp.
  \bibinfo{pages}{1086--1086}.
\bibitem[{Tan et~al.(2010)Tan, Viard-Gaudin, and Kot}]{Tan:2010}
\bibinfo{author}{G.~Tan}, \bibinfo{author}{C.~Viard-Gaudin},
  \bibinfo{author}{A.~Kot},
\newblock \bibinfo{title}{Individuality of alphabet knowledge in online writer
  identification},
\newblock \bibinfo{journal}{IJDAR} \bibinfo{volume}{13} (\bibinfo{year}{2010})
  \bibinfo{pages}{147--157}.
\bibitem[{Christlein et~al.(2017)Christlein, Bernecker, Hnig, Maier, and
  Angelopoulou}]{Christlein:2017}
\bibinfo{author}{V.~Christlein}, \bibinfo{author}{D.~Bernecker},
  \bibinfo{author}{F.~Hnig}, \bibinfo{author}{A.~Maier},
  \bibinfo{author}{E.~Angelopoulou},
\newblock \bibinfo{title}{Writer identification using gmm supervectors and
  exemplar-svms},
\newblock \bibinfo{journal}{Pattern Recogn.} \bibinfo{volume}{63}
  (\bibinfo{year}{2017}) \bibinfo{pages}{258–267}.
\bibitem[{Tang and Wu(2016)}]{Tang:2016}
\bibinfo{author}{Y.~Tang}, \bibinfo{author}{X.~Wu},
\newblock \bibinfo{title}{Text-independent writer identification via cnn
  features and joint bayesian},
\newblock in: \bibinfo{booktitle}{2016 15th International Conference on
  Frontiers in Handwriting Recognition (ICFHR)}, \bibinfo{year}{2016}, pp.
  \bibinfo{pages}{566--571}.
\bibitem[{Dosovitskiy et~al.(2014)Dosovitskiy, Springenberg, Riedmiller, and
  Brox}]{dosovitskiy:2014}
\bibinfo{author}{A.~Dosovitskiy}, \bibinfo{author}{J.~T. Springenberg},
  \bibinfo{author}{M.~Riedmiller}, \bibinfo{author}{T.~Brox},
\newblock \bibinfo{title}{Discriminative unsupervised feature learning with
  convolutional neural networks},
\newblock \bibinfo{journal}{Advances in neural information processing systems}
  \bibinfo{volume}{27} (\bibinfo{year}{2014}).
\bibitem[{Cohen et~al.(2017)Cohen, Afshar, Tapson, and van Schaik}]{Cohen:2017}
\bibinfo{author}{G.~Cohen}, \bibinfo{author}{S.~Afshar},
  \bibinfo{author}{J.~Tapson}, \bibinfo{author}{A.~van Schaik},
\newblock \bibinfo{title}{Emnist: Extending mnist to handwritten letters},
\newblock in: \bibinfo{booktitle}{2017 International Joint Conference on Neural
  Networks (IJCNN)}, \bibinfo{year}{2017}, pp. \bibinfo{pages}{2921--2926}.
\bibitem[{Springenberg et~al.(2014)Springenberg, Dosovitskiy, Brox, and
  Riedmiller}]{Springenberg:2014}
\bibinfo{author}{J.~Springenberg}, \bibinfo{author}{A.~Dosovitskiy},
  \bibinfo{author}{T.~Brox}, \bibinfo{author}{M.~Riedmiller},
\newblock \bibinfo{title}{Striving for simplicity: The all convolutional net}
  (\bibinfo{year}{2014}).
\bibitem[{Dalal and Triggs(2005)}]{Dalal:2005}
\bibinfo{author}{N.~Dalal}, \bibinfo{author}{B.~Triggs},
\newblock \bibinfo{title}{Histograms of oriented gradients for human
  detection},
\newblock in: \bibinfo{booktitle}{2005 IEEE Computer Society Conference on
  Computer Vision and Pattern Recognition (CVPR'05)},
  volume~\bibinfo{volume}{1}, \bibinfo{year}{2005}, pp.
  \bibinfo{pages}{886--893 vol. 1}.
\bibitem[{Zou et~al.(2004)Zou, Hastie, and Tibshirani}]{Zou04}
\bibinfo{author}{H.~Zou}, \bibinfo{author}{T.~Hastie},
  \bibinfo{author}{R.~Tibshirani},
\newblock \bibinfo{title}{Sparse principal component analysis},
\newblock \bibinfo{journal}{Journal of Computational and Graphical Statistics}
  \bibinfo{volume}{15} (\bibinfo{year}{2004}) \bibinfo{pages}{2006}.
\bibitem[{Marti and Bunke(2002)}]{IAM}
\bibinfo{author}{U.-V. Marti}, \bibinfo{author}{H.~Bunke},
\newblock \bibinfo{title}{The iam-database: An english sentence database for
  offline handwriting recognition},
\newblock \bibinfo{journal}{International Journal on Document Analysis and
  Recognition} \bibinfo{volume}{5} (\bibinfo{year}{2002})
  \bibinfo{pages}{39--46}.
\bibitem[{Burges(1998)}]{Burges:1998}
\bibinfo{author}{C.~J.~C. Burges},
\newblock \bibinfo{title}{A tutorial on support vector machines for pattern
  recognition} \bibinfo{volume}{2} (\bibinfo{year}{1998})
  \bibinfo{pages}{121–167}.
\bibitem[{Kleber et~al.(2013)Kleber, Fiel, Diem, and Sablatnig}]{CVL}
\bibinfo{author}{F.~Kleber}, \bibinfo{author}{S.~Fiel},
  \bibinfo{author}{M.~Diem}, \bibinfo{author}{R.~Sablatnig},
\newblock \bibinfo{title}{Cvl-database: An off-line database for writer
  retrieval, writer identification and word spotting},
\newblock in: \bibinfo{booktitle}{2013 12th International Conference on
  Document Analysis and Recognition}, \bibinfo{year}{2013}, pp.
  \bibinfo{pages}{560--564}.
\bibitem[{Brink et~al.(2012)Brink, Smit, Bulacu, and Schomaker}]{Brink:2012}
\bibinfo{author}{A.~Brink}, \bibinfo{author}{J.~Smit},
  \bibinfo{author}{M.~Bulacu}, \bibinfo{author}{L.~Schomaker},
\newblock \bibinfo{title}{Writer identification using directional ink-trace
  width measurements},
\newblock \bibinfo{journal}{Pattern Recognition} \bibinfo{volume}{45}
  (\bibinfo{year}{2012}) \bibinfo{pages}{162--171}.
\bibitem[{Sheng and Schomaker(2017)}]{Sheng:2017}
\bibinfo{author}{H.~Sheng}, \bibinfo{author}{L.~Schomaker},
\newblock \bibinfo{title}{Beyond ocr: Multi-faceted understanding of
  handwritten document characteristics},
\newblock \bibinfo{journal}{Pattern Recognition} \bibinfo{volume}{63}
  (\bibinfo{year}{2017}) \bibinfo{pages}{321--333}.
  \DOIprefix\doi{10.1016/j.patcog.2016.09.017}.
\bibitem[{Sheng and Schomaker(2016)}]{sheng:2017C}
\bibinfo{author}{H.~Sheng}, \bibinfo{author}{L.~Schomaker},
\newblock \bibinfo{title}{Writer identification using curvature-free features},
\newblock \bibinfo{journal}{Pattern Recognition} \bibinfo{volume}{63}
  (\bibinfo{year}{2016}) \bibinfo{pages}{451–464}.
\bibitem[{Nguyen et~al.(2018)Nguyen, Nguyen, Ino, Indurkhya, and
  Nakagawa}]{Nguyen:2018}
\bibinfo{author}{H.~Nguyen}, \bibinfo{author}{C.~Nguyen},
  \bibinfo{author}{T.~Ino}, \bibinfo{author}{B.~Indurkhya},
  \bibinfo{author}{M.~Nakagawa},
\newblock \bibinfo{title}{Text-independent writer identification using
  convolutional neural network},
\newblock \bibinfo{journal}{Pattern Recognition Letters} \bibinfo{volume}{121}
  (\bibinfo{year}{2018}).
\bibitem[{Khalifa et~al.(2015)Khalifa, Al-Maadeed, Tahir, Bouridane, and
  Jamshed}]{Khalifa:2015}
\bibinfo{author}{E.~Khalifa}, \bibinfo{author}{S.~A. Al-Maadeed},
  \bibinfo{author}{M.~A. Tahir}, \bibinfo{author}{A.~Bouridane},
  \bibinfo{author}{A.~Jamshed},
\newblock \bibinfo{title}{Off-line writer identification using an ensemble of
  grapheme codebook features},
\newblock \bibinfo{journal}{Pattern Recognit. Lett.} \bibinfo{volume}{59}
  (\bibinfo{year}{2015}) \bibinfo{pages}{18--25}.
\bibitem[{Hadjadji and Chibani(2018)}]{Hadjadji:2018}
\bibinfo{author}{B.~Hadjadji}, \bibinfo{author}{Y.~Chibani},
\newblock \bibinfo{title}{Two combination stages of clustered one-class
  classifiers for writer identification from text fragments},
\newblock \bibinfo{journal}{Pattern Recognition} \bibinfo{volume}{82}
  (\bibinfo{year}{2018}). \DOIprefix\doi{10.1016/j.patcog.2018.05.001}.

\end{thebibliography}

\end{document}